\pdfoutput=1

\documentclass[11pt]{article}

\usepackage{acl}

\usepackage{times}
\usepackage{latexsym}

\usepackage[T1]{fontenc}

\usepackage[utf8]{inputenc}
\usepackage{microtype}
\usepackage[most]{tcolorbox}
\usepackage{algorithm}
\usepackage{anyfontsize}
\usepackage{dirtytalk}
\usepackage{geometry}
\usepackage{booktabs}
\usepackage{amsmath}
\usepackage[normalem]{ulem}
\useunder{\uline}{\ul}{}
\usepackage{booktabs}  
\usepackage{graphicx} 
\usepackage{listings}
\usepackage[wby]{callouts}
\usepackage{t1enc}
\usepackage{booktabs}
\usepackage{times}
\usepackage{multirow}
\usepackage{graphicx}
\usepackage{amsmath}
\usepackage{algorithm}
\usepackage[noend]{algpseudocode}
\usepackage{diagbox}
\usepackage{latexsym}
\usepackage{enumitem}
\usepackage{siunitx}

\usepackage{helvet}  
\usepackage{courier}  
\usepackage{url}  
\usepackage{graphicx}  
\usepackage{courier}
\usepackage{amsmath}
\usepackage[normalem]{ulem}
\useunder{\uline}{\ul}{}
\usepackage{graphicx}
\usepackage{multirow}
\usepackage{mathbbol}
\usepackage{caption}
\usepackage{CJKutf8}
\usepackage{verbatim}
\usepackage{float}
\usepackage{booktabs}
\usepackage{array}
\usepackage{times}
\usepackage{csquotes}
\usepackage{latexsym}
\usepackage{tablefootnote}
\usepackage[normalem]{ulem}
\usepackage{todonotes}
\usepackage{subcaption}
\usepackage{titlesec}

\usepackage{microtype}

\newcommand{\mypar}[1]{\noindent{\textbf{#1\ }}}
\titlespacing*{\section}
{0pt}{.5ex plus .5ex minus .2ex}{.5ex plus .2ex minus .2ex}
\titlespacing*{\subsection}
{0pt}{.5ex plus .5ex minus .2ex}{.5ex plus .2ex minus .2ex}

\title{\BaseName: Generating Puns with Ambiguous Context}
\author{Anirudh Mittal\textsuperscript{2}\thanks{~~Equal contribution.} ~\thanks{~~Work done when the author is interning at UCLA.},
Yufei Tian\textsuperscript{1*},
  Nanyun Peng\textsuperscript{1} \\
 \textsuperscript{1} Computer Science Department, University of California, Los Angeles, \\
   \textsuperscript{2} Centre for Machine Intelligence and Data Science, Indian Institute of Technology Bombay\\
   \texttt{ anirudhmittaljpr@gmail.com},
  \texttt{ \{yufeit, violetpeng\}@cs.ucla.edu} \\}

\begin{document}
\maketitle
\begin{abstract}
We propose a simple yet effective way to generate pun sentences that does not require any training on existing puns. Our approach is inspired by humor theories that ambiguity comes from the context rather than the pun word itself. Given a pair of definitions of a pun word,\footnote{We focus on generating homographic puns where two or more meanings of a word form an intended humorous effect.} our model first produces a list of related concepts through a reverse dictionary to identify unambiguous words to represent the pun and the alternative senses. We then utilize one-shot GPT3 to generate context words and then generate puns incorporating context words from both senses.
Human evaluation shows that our method successfully generates puns 52\% of the time, outperforming  well crafted baselines and the state-of-the-art models by a large margin. 
\end{abstract}

\section{Introduction}

\begin{figure}[t!]
  \centering
    \includegraphics[width=0.45\textwidth]{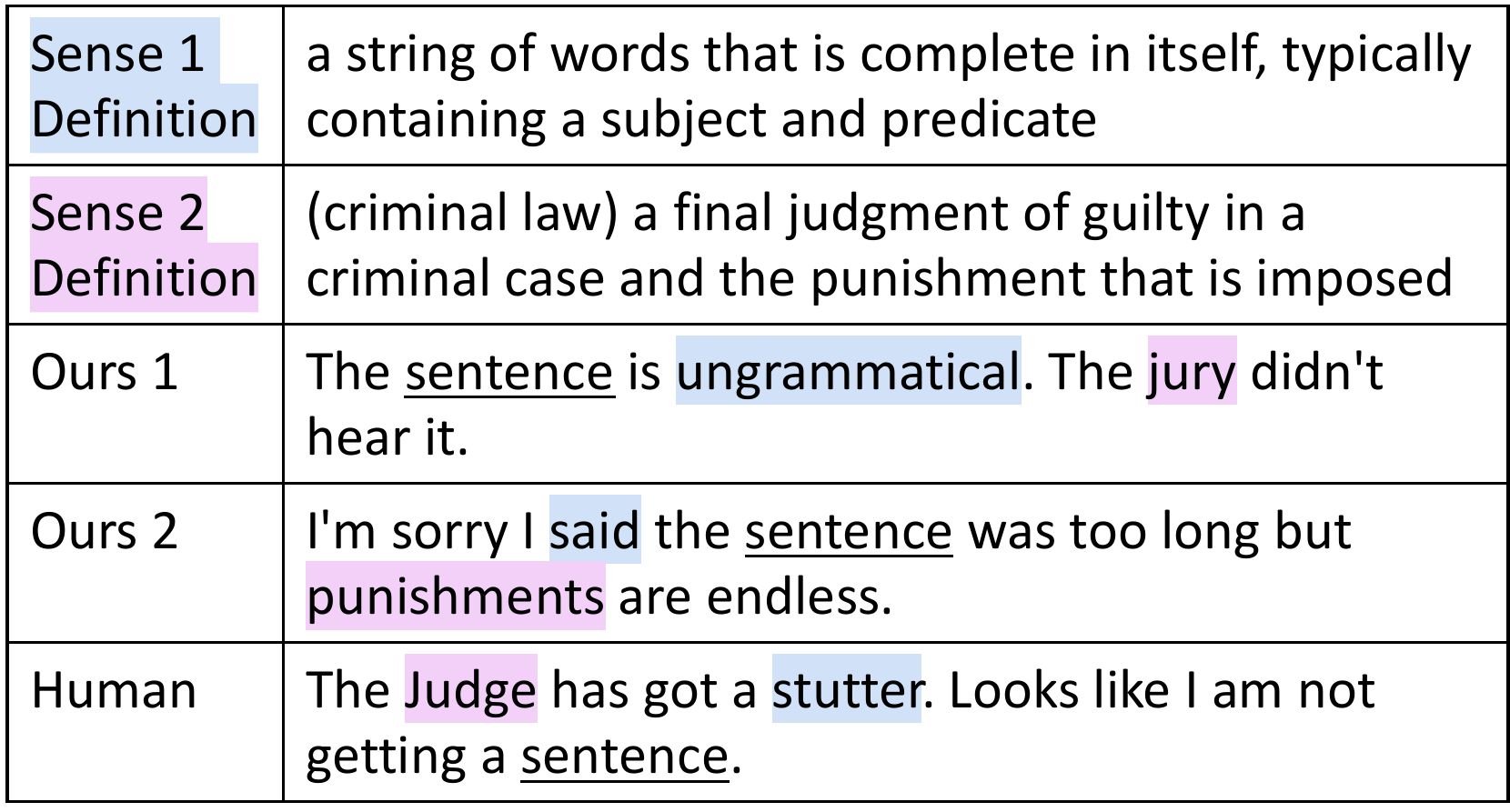}
  \caption{An illustration of homographic puns. The target pun word \textit{`sentence'} and the two sense definitions are given as input. To make the target word interpretable in both senses, we propose to include context words (highlighted in blue and pink) related to both senses.}
  \label{fig:illustration}    
  \vspace{-1.7em}
\end{figure}

Computational humor has garnered interest in the NLP community~\cite{petrovic-matthews-2013-unsupervised,miller2017semeval,zou-lu-2019-joint,garimella-etal-2020-judge,yang-etal-2020-textbrewer}.
In this paper, we tackle the problem of generating homographic puns \cite{miller2017semeval}: two or more meanings of a word form an intended humorous effect. For example, the three puns listed in Figure \ref{fig:illustration} exploit two contrasting meanings of the word \textit{sentence}: 1) a grammatical string of words and 2) the punishment by a court assigned to a guilty person. 

Due to the lack of sizeable training data, existing approaches are heavy-weighted in order to \textit{not} rely on pun sentences for training. For example, \cite{yu2018neural} train a constrained neural language model \cite{mou2015backward} from a general text corpus and then use a joint decoding algorithm to promote puns. \citet{he2019pun} propose a local-global surprisal principle, and \citet{luo2019pun} leverage the Generative Adversarial Nets \cite{goodfellow2014generative} to encourage ambiguity of the outputs via reinforcement learning. We, on the other hand, propose a simple yet effective way to tackle this problem: \textit{encouraging ambiguity by incorporating context words related to each sense}. 

Inspired by humor theories \cite{lippman2000contextual}, we hypothesize that it is the contextual connections rather than the pun word itself that are crucial for the success of pun generation. For instance, in Figure \ref{fig:illustration} we observe that context related to both senses (e.g., \textit{ungrammatical} and \textit{jury}) appear in a punning sentence. Such observation is important as the error analysis of the SOTA model \cite{luo2019pun} shows that 46\% of the outputs fail to be puns due to single word sense, and another 27\% fail due to being too general, both of which can be resolved by introducing more context.

\begin{figure*}[ht!]
  \centering
    \includegraphics[width=0.9\textwidth]{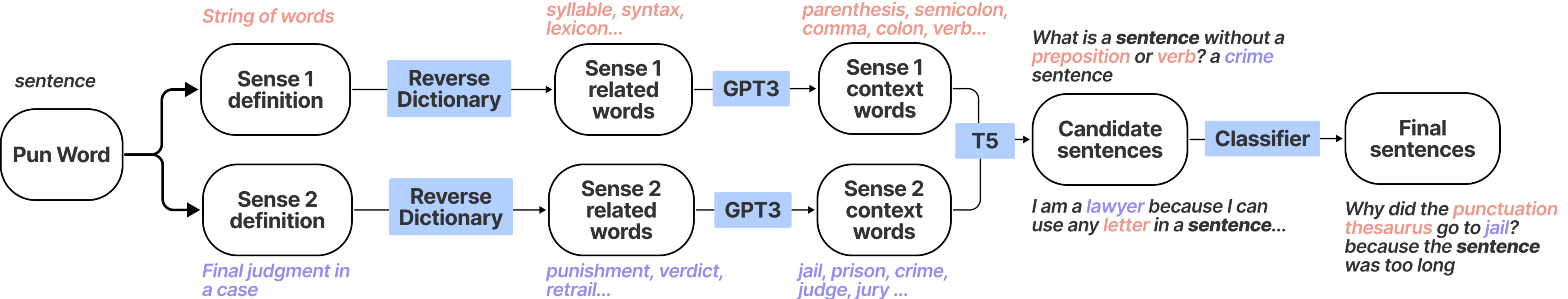}
  \caption{
  Overview of the approach. We also give an example for pun word `sentence' for each step.} 
  
  \label{fig:overview}    
  \vspace{-1.7em}
\end{figure*}

Specifically, given the two sense definitions of a target pun word, we first use a reverse dictionary to generate \textbf{related words} that are monosemous for both senses. This first step helps us circumvent the obstacle of processing pun words with the same written form. 
We then propose to use \textbf{context words} to link the contrasting senses and make our target pun word reasonable when interpreted in both definitions. We explore three different settings: extractive-based, similarity-based, and generative-based. Finally, we finetune the T5 model \cite{raffel2020exploring} on general non-humorous texts to generate coherent sentences given the pun word and contexts words as input. 

Our experimental results show that retrieve-and-extract context words outperforms the giant few-shot GPT3 model in terms of generating funny pun sentences, although the latter has shown to be much more powerful in many sophisticated tasks \cite{brown2020language}.  Our simple pipeline remarkably outperforms all of the more heavy-weighted approaches. Our code and data is available at \url{https://github.com/PlusLabNLP/AmbiPun}.

\section{Methodology} \label{sec:method}
\mypar{Overview and Motivation} 
Our input is the target pun word ($p$) and its two sense definitions ($S_1$, $S_2$), and the output is a \textit{list} of humorous punning sentences. 
We implement the ambiguity principle proposed in~\citep{kao2016computational}: a pun sentence should contain one or more context words corresponding to each of the two senses.\footnote{Note that all previous works produce only the \textit{best} sentence during decoding time, while we aim at generating \textit{tens or hundreds} of sentences for a target pun word so that our task is actually more challenging.}
The overview of our approach is visualized in Figure \ref{fig:overview}. 
 
Given $S_1$ and $S_2$, we first use a reverse dictionary to generate a list of words that semantically match the query descriptions. We call them \textbf{related words} (Section \ref{subsec:related-words}). However, those related words are synonyms of the pun word and are rarely observed as it is in humorous sentences. For example, for the sentence: ``The Judge has got a stutter. Looks like I am not getting a sentence.'', The word representing the first sense (i.e. a final judgment of guilty in a criminal case)  is represented by \textit{Judge}, which could not be generated using the sense definition.
 
 Considering such nuances, in Section \ref{subsec:context} we propose three different methods to obtain the \textbf{context words}
 . They are extractive, similarity, and generative based. Finally, in Section \ref{subsec:candidate-gen} and \ref{subsec:classification}, we introduce a keyword-to-text generator to generate candidate sentences 
 , and a humor classifier to rule out some of the non-pun sentences. Final sentences 
 are then randomly sampled for evaluation. All our training data is general, non-humorous corpus except for the humor classifier.

\subsection{Generating Related Words} \label{subsec:related-words}
We aim at differentiating the two senses of a polysemy by taking the related words, so that each sense will be represented by a set of monosemous words. To this end, we leverage the Reverse Dictionary \cite{qi2020wantwords, zhang2020multi} which takes as input a description and generates multiple related words whose semantic meaning match the query description. For each sense definition, we generate five words.

\subsection{Generating Context Words} \label{subsec:context}
For context words, we compare three different approaches. As an example, we compare the output of context words for the pun word `sentence' in Table \ref{table: context words} in the appendix. Refinement of the context words is mentioned in section \ref{subsec:implement} in the appendix.

\mypar{Method 1: Extractive (TF-IDF)}
For each related word, we retrieve sentences from the One Billion Word dataset containing that word and then extract keywords using RAKE \cite{rose2010automatic} from the retrieved sentences. 
Based on this TF-IDF value, we choose the top 10 context words that are mostly likely to be used along with the related words and therefore the pun word. 

\mypar{Method 2: Similarity (Word2Vec)}
Inspired by the idea that ``a word is characterized by the company it keeps'', we propose to get context words from word2vec. 
\cite{ghazvininejad2016generating} have also shown that the training corpus for word2vec plays a crucial role on the quality of generated context words. Hence, we train on Wikipedia which has a comprehensive coverage of diverse topics.

\mypar{Method 3: Generative (Few-shot GPT3)}
For the generative version, we provide the powerful language model GPT3 \citep{brown2020language} with two examples and generate context words. Details about the prompt can be found in section \ref{sec:gpt3 appendix} of the appendix.

\subsection{Candidate Sentence generation} \label{subsec:candidate-gen}

After receiving context words for each sense, we generate humorous puns. We finetune a keyword-to-sentence model using T5 \citep{raffel2020exploring}, as it is capable of handling text-to-text tasks. 
The prompt provides the pun word, and two context words from each of the two senses. 
For example for the word `sentence', a possible prompt can be \textit{generate sentence: sentence, judge, trial, noun, comma.} Moreover, we also investigate whether the position of the pun word will affect the quality of generated sentences.  We insert the pun word in the first, third, and fifth place of the prompt. We discuss our findings in Section \ref{sec:position}.

\subsection{Humor Classification} \label{subsec:classification}
Finally, we introduce a classification model to assist us in selecting (i.e., ranking) punning sentences. 
Since we do not have sizable training data for puns, we propose to train our classification model on humorous dataset in a distantly supervised fashion.  
Specifically, we train BERT-large \citep{devlin2018bert} on the ColBERT dataset \citep{annamoradnejad2020colbert} that contains 200,000 jokes and non-jokes used for humor detection. We use the probability produced by the classification model to rank our candidate sentences. 

Our error analysis in section Appendix.\ref{appendix:humor-classification} shows that our classifier can successfully rule out the bad generations, i.e., non-puns, as puns are humorous by nature. However, the model is not great at choosing the best samples. Therefore, we use this classifier only to remove the bottom third candidates. We leave this for open future work to accurately pick out high-quality punning sentences instead of funny sentences. 
\section{Experiments}

\subsection{Datasets}

\mypar{Training:}
For the context word generation step, we use the One Billion word dataset \citep{chelba2013one} to retrieve sentences for a given word and calculate TF-IDF values. This dataset contains roughly 0.8B words and is obtained from WMT 2011 News crawl data. For the humor classifier and candidate generation module, we use ColBERT dataset \citep{annamoradnejad2020colbert}. It contains 100k jokes and 100k non-jokes.

\mypar{Evaluation dataset:}
Following other pun generation works, we use the SemEval 2017 Task 7 \citep{miller2017semeval} for evaluation. The dataset contains 1,163 human written punning jokes with a total of 895 unique pun words. Each sentence has the target pun word, location of the pun word and the WordNet sense keys of the two senses. 
\subsection{Baselines}
\label{subsec: baselines}

\mypar{Neural Pun} \citet{yu2018neural} propose the first neural approach to homographic puns based on a constrained beam search algorithm to jointly decode the two distinct senses of the same word.

\mypar{Pun-GAN} The SOTA introduced by \citet{luo2019pun} that adopts the Generative Adversarial Net to generate homographic puns.

\mypar{Few-shot GPT3} We generate puns with a few examples feeding into GPT3 \textit{davinci-instruct-beta}, the most capable model in the GPT3 family for creative generation. 
We provide the target pun word and its two senses in our prompt, along with the instruction to generate puns.

\mypar{Ablations of our own models} We also compare three methods proposed by us to obtain the context words (described in Section \ref{subsec:context}). We call them \ModelName, \VOneName, and \VTwoName. 

\begin{table}[t]
\small
\centering
\renewcommand{\tabcolsep}{1.6mm}
\begin{tabular}{@{}l|c|cc|cc@{}}
\toprule
\multicolumn{1}{c|}{\multirow{2}{*}{\textbf{Model}}} &
  \multicolumn{1}{c|}{\small{\multirow{2}{*}{\begin{tabular}[c]{@{}c@{}}\textbf{Avg}\\ \textbf{Seq}\\ \textbf{Len}\end{tabular}}}} &
  \multicolumn{2}{c|}{\textbf{Corpus-Div}} &
  \multicolumn{2}{c}{\textbf{Sentence-Div}}   \\
 \cmidrule(lr){3-6}
 &
  \multicolumn{1}{c|}{} &
  \multicolumn{1}{c}{Dist-1} &
  \multicolumn{1}{c|}{Dist-2} &
  \multicolumn{1}{c}{Dist-1} &
  \multicolumn{1}{c}{Dist-2}
   \\ \midrule
Few-shot  GPT3 & 12.3 & \textbf{37.1} & \textbf{80.4} & {\ul 94.5}    & {\ul 85.7}  \\
Neural   Pun    & 12.6 & 30.2          & 73.0          & 91.3          & 90.5         \\
Pun GAN         & 9.7  & {\ul 34.6}          & 71.9          & 90.2          & 87.6     \\ \midrule
\VOneName        & 13.4 & 32.4          & 77.1          & 92.9          & 91.2  \\
\VTwoName        & {\ul 13.5} & 32.8    & 77.8          & 93.6          & 91.2  \\
\ModelName       & \textbf{14.0} & 31.7          & {\ul 78.7}    & \textbf{96.3} & \textbf{92.3}\\ \midrule
Human           & 14.1 & 36.6          & 81.9          & 95.5          & 92.4        \\ \bottomrule
\end{tabular}
\caption{Results of automatic evaluation on average sequence length, sentence-level and corpus-level diversity. Boldface denotes the best performance and underline denotes the second best performance among systems.}
\label{table: automatic evaluation}
\vspace{-1em}
\end{table}

\begin{table}[t]
\small
\centering
\begin{tabular}{@{}lccc@{}}
\toprule
Model &
  \begin{tabular}[c]{@{}c@{}}Success\\ Rate\end{tabular} &
  \begin{tabular}[c]{@{}l@{}}Fun\end{tabular} &
  \begin{tabular}[c]{@{}l@{}}Coherence\end{tabular} \\ \midrule
Few-shot   GPT3 & 13.0\%           & 1.82          & \textbf{3.77} \\
Neural   Pun    & 35.3\%           & 2.17          & 3.21          \\
Pun GAN         & 35.8\%           & 2.28          & 2.97          \\ \midrule
\VOneName        & 45.5\%           & 2.69          & 3.38          \\
\VTwoName        & {\ul 50.5\%}     & {\ul 2.94}    & {\ul 3.53}    \\
\ModelName       & \textbf{  52.2\%}*  & \textbf{ 3.00}* & 3.48          \\ \midrule
Human           & 70.2\%           & 3.43          & 3.66          \\ \bottomrule
\end{tabular}
\caption{Human evaluation results on all the pun generation systems. We show the success rates, and average scores of funniness and coherence. Overall, \ModelName{} performs the best. The superiority of our model in terms of success rate and funniness is statistically significant over the best baseline and is marked by *.}
\label{table:human evaluation}
\vspace{-1.5em}
\end{table}

 \begin{table*}[t!]
\centering
\fontsize{9}{8}\selectfont {
\begin{tabular}{l| ccc}
\toprule
\textbf{Pun word}    & \multicolumn{3}{l}{\textbf{Irrational}}                                                                                                       \\  
Sense 1     & \multicolumn{3}{l}{Real but not expressible as the quotient of two integers}                                                         \\ 
Sense 2     & \multicolumn{3}{l}{Not consistent with or using reason}                                                                              \\ \midrule
\textbf{Model}       & \multicolumn{1}{c|}{\textbf{Example}}                                                             &     \multicolumn{1}{c|}{\textbf{Pun}} & \textbf{Funny} \\ \hline
GPT3        & \multicolumn{1}{l|}{I can't make a decision with all this irrationality going on.}                                                                     & \multicolumn{1}{c|}{No}   & 1.4          \\ 
Neural Pun  & \multicolumn{1}{l|}{Note that this means that there is an irrational problem.}                & \multicolumn{1}{c|}{Yes}   & 2.4           \\ 
Pun-GAN     & \multicolumn{1}{l|}{It can be use the irrational system.}                                     & \multicolumn{1}{c|}{No}   & 1.2           \\ 
\ModelName & \multicolumn{1}{l|}{I have an irrational \uwave{paranoia} about \uline{mathematical}  \uline{integers}.}    & \multicolumn{1}{c|}{Yes}   & \textbf{3.8}           \\ 
\VTwoName  & \multicolumn{1}{l|}{My \uline{calculator} is unjust and \uwave{illogic}. It's irrational.} & \multicolumn{1}{c|}{Yes}   & 3.4           \\ 
Human       & \multicolumn{1}{l|}{Old math teachers never die, they just become irrational.}               & \multicolumn{1}{c|}{Yes}   & \textbf{3.8}           \\ \bottomrule
\toprule
\textbf{Pun word}    & \multicolumn{3}{l}{\textbf{Drive}}                                                                                                          \\ 
Sense 1     & \multicolumn{3}{l}{A journey in a vehicle (usually an automobile)}                                                    \\ 
Sense 2 &
  \multicolumn{3}{l}{\begin{tabular}[c]{@{}c@{}}The trait of being highly motivated\end{tabular}} \\ \midrule
\textbf{Model}       & \multicolumn{1}{c|}{\textbf{Example}}                                                                 & \multicolumn{1}{c|}{\textbf{Pun}} & \textbf{Funny} \\
GPT3        & \multicolumn{1}{l|}{I am exhausted, I need a nap before I can drive any more.}                                                                     & \multicolumn{1}{c|}{No}   & 2.0           \\ 
Neural Pun  & \multicolumn{1}{l|}{It is that it can be use to drive a variety of function?}                     & \multicolumn{1}{c|}{No}   & 1.6           \\ 
Pun-GAN     & \multicolumn{1}{l|}{In he drive to the first three years.}                                            & \multicolumn{1}{c|}{No}   & 1.2           \\ 
\ModelName &
  \multicolumn{1}{l|}{\begin{tabular}[c]{@{}l@{}}What do you call a \uwave{genius} with cunning drive? \uline{racecar} \uline{driver}. \end{tabular}} &
  \multicolumn{1}{c|}{Yes} &
  3.6 \\ 
\VTwoName  & \multicolumn{1}{l|}{I have the \uline{determination} to \uwave{travel} to my \uwave{destination}. But i don't have the drive.}         & \multicolumn{1}{c|}{Yes}   & 4.0           \\ 
Human       & \multicolumn{1}{l|}{A boy saving up for a car has a lot of drive.}                         & \multicolumn{1}{c|}{Yes}   & \textbf{4.2}           \\ \bottomrule

\end{tabular}
\caption{Generated sentences for the word `Irrational' and `Drive'and their sense definitions. We underline the context words that are related to each sense. All the generations are evaluated by external annotators, not the authors.}
\label{table: case study}
\vspace{-1.7em}}
\end{table*}

\subsection{Evaluation} \label{subsec: evaluation}
\mypar{Automatic Evaluation}
 We follow \citet{luo2019pun, yu2018neural} to calculate distinct unigram and bigrams as the \textit{diversity} \cite{li2016diversity} in terms of sentence level and corpus level. We also report the the average sentence length produced.

\mypar{Human Evaluation} 
We randomly select 100 sentences and collected our human ratings on Amazon Mechanical Turk (AMT). For each sentence, three workers are explicitly given the target pun word and its two sense definitions provided by the Sem\-Eval 2017 Task 7. We first ask them to judge if a given sentence is a successful pun sentence. Then, they are asked to rate the funniness and coherence on a scale from 1 (not at all) to 5 (extremely). Besides detailed instructions and explanation for each criteria, we also adopt attention questions and qualification types to rule out irresponsible workers. We conduct paired t-test for significance testing. The difference between our best performing model and the best baseline is significant.

\section{Results and Analysis}
\subsection{Pun Generation Results}
\mypar{Automatic Evaluation}
Results of the automatic evaluation can be seen in Table \ref{table: automatic evaluation}. The average sentence length of our model is closest to human and gets the highest sentence-level diversity. Although our baseline Pun-GAN and Few-shot GPT3 have higher distinct unigram ratios at the corpus level, that is because all baseline models generate \textit{one} sentence per pun word, while \BaseName{} generates \textit{tens} of sentences per pun word, which inevitably sacrifices topical diversity. Nevertheless, \BaseName{} achieves the highest corpus-level bi-gram diversity.

\mypar{Human Evaluation} Results from the automatic evaluation can be seen in Table~\ref{table:human evaluation}. We evaluate the success rate, funniness, and coherence of the generated outputs. The superiority of our models are obvious. For significance testing, we conducted paired t-test, where our systems outperformed the best baseline in terms of success rate and funniness (p-value < 0.05). On the other hand, GPT3 could generate even more coherently than humans. 

\mypar{Analysis between extractive and generative method.} Interestingly, the extractive method has higher success rates (p-value < 0.05) and is funnier (p-value < 0.07) than the generative method, indicating that extracting salient words from human written sentences could introduce more surprising and uncommon words than language models. We posit that those atypical words refresh people's eyes and thus boost the pun success rate as well as the funniness score. On the other hand, we also tried to equip GPT3 with greater creatively by top-k sampling with a large temperature $T$. However, larger $T$s also result in arbitrary responses that humans may find unreadable.  We hope our discovery could draw the community's attention to those traditional techniques for creative generation.

\subsection{Case Study}

To better understand the advantages of our method, we conduct a case study for the pun word ``Irrational'' and ``Drive'' in Table \ref{table: case study}. For both target pun words, at most one of the baselines successfully generates a punning sentence. As discussed earlier, one possible reason is the absence of both senses. 
On the other hand, both \ModelName \space and \VOneName \space introduce context words for the two senses and thus are able to generate of high quality puns that almost match the human written puns in terms of the funniness score.

\section{The Position of Pun Words} \label{sec:position}

As is mentioned in Section \ref{subsec:candidate-gen}, we play with the position of the pun word in the prompt given to the candidate sentence generation model. We try three variants by putting the target pun word at the start, in the middle, and at the end. For each variant, we ask Mechanical Turkers to judge if the given sentences are puns. Again, each sentence is rated by three Turkers and we take the majority answer if the workers disagree. Results from this analysis can be seen in Table \ref{table: position}. We observe that people find a sentence more likely to be a pun when the target word appears at the end.   

To verify such hypothesis, we also calculate the position of the pun words of 1,163 human written pun sentences from SemEval 2017 Task 7 and report the distribution in Figure \ref{fig: position} in the Appendix. The histogram corroborates with the human annotated samples in that both suggest that keeping the pun word at the end of the sentence generates funnier puns. Theory of humor which says that the "joke" in a funny sentences some towards the end of the sentence \cite{shahaf2015inside} validates our analysis. 

\begin{table}[t]
\centering
\small
\begin{tabular}{l c }
\hline
                   & Success Rate \\ \hline
Beginning & 46.7\%                  \\
Middle    & 52.0\%                  \\ 
End       & 54.7\%                  \\ \hline
\end{tabular}
\caption{The pun success rate sentences based on their position annotated by human.}
\label{table: position}
\vspace{-1.5em}
\end{table}

\section{Related Works}


\subsection{Creative Text Generation}
\paragraph{Pun generation.}
Many of the previous works on pun generation have focused on phonological or syntactic pattern rather than semantic pattern \cite{miller-gurevych-2015-automatic, hong-ong-2009-automatically, petrovic-matthews-2013-unsupervised,inproceedings2000} thus lacking creativity and flexibility.  \citet{he2019pun} make use of local-global surprisal principle to generate homophonic puns and \citet{yu-etal-2020-homophonic} uses constrained lexical rewriting for the same task. \citet{hashimoto2018retrieveandedit} use a retrieve and edit approach to generate homographic puns and \citet{yu2018neural,luo2019pun} propose complex neural model architecture such as constrained language model and GAN, and do not put emphasis on the linguistic structure of puns. We identify their absence of both the senses as a shortcoming and build our approach from there. 

\paragraph{Humor generation.} 
Humor generation still remains an unsolved problem, and is usually studied in a specific setting.~\citet{petrovic-matthews-2013-unsupervised} generates joke of the type `I like my X like I like my Y, Z'. \citet{garimella-etal-2020-judge} develops a model to fill blanks in madlibs format and \citet{yang-etal-2020-textbrewer} edit headlines to make them funny. More research is required to generate humorous sentences that are not constrained by their semantic structure.

\paragraph{Figurative language generation.} In addition to pun, there are many attempts to generate figurative language such as metaphor, simile~\cite{chakrabarty2020generating}, sarcasm, etc.
\citet{yu-wan-2019-avoid} use metaphorically used verbs to generate metaphors in an unsupervised fashion, while \citet{chakrabarty2021mermaid,stowe2021metaphor} generate metaphors using symbolism and discriminative and conceptual mapping. \citet{mishra2019modular} propose a modular architecture for unsupervised sarcasm generation, and \citet{chakrabarty2020r} use commonsense knowledge for the same task. \citet{tian-etal-2021-hypogen-hyperbole} on the other hand are the first leverage semantic structure and commonsense and counterfactual knowledge to generate hyperboles.

\subsection{Pun detection} SemEval 2017 Task 7 \cite{miller2017semeval} introduced the challenge of pun detection, location detection and sense interpretation for homographic and homophonic puns.~\citet{10.1145/3308558.3313505} make use of Gated Attention network to detection homophonic puns.~\citet{zou-lu-2019-joint} introduces a tagging schemes which lets them detect puns as well as their location. They apply this approach to both homophonic and homographic puns.

\section{Conclusion}

We propose a novel approach towards homographic puns generation. Unlike previous works that are mathematically heavy, our approach is backed by the humor theory that ambiguity is achieved by the context. Automatic and human evaluations show that our model \BaseName{} outperforms the current state-of-the-art model by a large margin.
\section*{Acknowledgments}
The authors would like to thank the members of PLUSLab and the anonymous reviewers for helpful comments. Yufei Tian is supported by an Amazon Fellowship.

\bibliography{anthology,custom}

\begin{thebibliography}{37}
\expandafter\ifx\csname natexlab\endcsname\relax\def\natexlab#1{#1}\fi

\bibitem[{Annamoradnejad and Zoghi(2020)}]{annamoradnejad2020colbert}
Issa Annamoradnejad and Gohar Zoghi. 2020.
\newblock Colbert: Using bert sentence embedding for humor detection.
\newblock \emph{arXiv preprint arXiv:2004.12765}.

\bibitem[{Braslavski et~al.(2018)Braslavski, Blinov, Bolotova, and
  Pertsova}]{braslavski2018evaluate}
Pavel Braslavski, Vladislav Blinov, Valeria Bolotova, and Katya Pertsova. 2018.
\newblock How to evaluate humorous response generation, seriously?
\newblock In \emph{Proceedings of the 2018 Conference on Human Information
  Interaction \& Retrieval}, pages 225--228.

\bibitem[{Brown et~al.(2020)Brown, Mann, Ryder, Subbiah, Kaplan, Dhariwal,
  Neelakantan, Shyam, Sastry, Askell et~al.}]{brown2020language}
Tom~B Brown, Benjamin Mann, Nick Ryder, Melanie Subbiah, Jared Kaplan, Prafulla
  Dhariwal, Arvind Neelakantan, Pranav Shyam, Girish Sastry, Amanda Askell,
  et~al. 2020.
\newblock Language models are few-shot learners.
\newblock \emph{arXiv preprint arXiv:2005.14165}.

\bibitem[{Chakrabarty et~al.(2020{\natexlab{a}})Chakrabarty, Ghosh, Muresan,
  and Peng}]{chakrabarty2020r}
Tuhin Chakrabarty, Debanjan Ghosh, Smaranda Muresan, and Nanyun Peng.
  2020{\natexlab{a}}.
\newblock R3: Reverse, retrieve, and rank for sarcasm generation with
  commonsense knowledge.
\newblock \emph{arXiv preprint arXiv:2004.13248}.

\bibitem[{Chakrabarty et~al.(2020{\natexlab{b}})Chakrabarty, Muresan, and
  Peng}]{chakrabarty2020generating}
Tuhin Chakrabarty, Smaranda Muresan, and Nanyun Peng. 2020{\natexlab{b}}.
\newblock Generating similes effortlessly like a pro: A style transfer approach
  for simile generation.
\newblock \emph{arXiv preprint arXiv:2009.08942}.

\bibitem[{Chakrabarty et~al.(2021)Chakrabarty, Zhang, Muresan, and
  Peng}]{chakrabarty2021mermaid}
Tuhin Chakrabarty, Xurui Zhang, Smaranda Muresan, and Nanyun Peng. 2021.
\newblock \href {http://arxiv.org/abs/2103.06779} {Mermaid: Metaphor generation
  with symbolism and discriminative decoding}.

\bibitem[{Chelba et~al.(2013)Chelba, Mikolov, Schuster, Ge, Brants, Koehn, and
  Robinson}]{chelba2013one}
Ciprian Chelba, Tomas Mikolov, Mike Schuster, Qi~Ge, Thorsten Brants, Phillipp
  Koehn, and Tony Robinson. 2013.
\newblock One billion word benchmark for measuring progress in statistical
  language modeling.
\newblock \emph{arXiv preprint arXiv:1312.3005}.

\bibitem[{Devlin et~al.(2018)Devlin, Chang, Lee, and
  Toutanova}]{devlin2018bert}
Jacob Devlin, Ming-Wei Chang, Kenton Lee, and Kristina Toutanova. 2018.
\newblock Bert: Pre-training of deep bidirectional transformers for language
  understanding.
\newblock \emph{arXiv preprint arXiv:1810.04805}.

\bibitem[{Diao et~al.(2019)Diao, Lin, Yang, Fan, Wu, Zhang, and
  Xu}]{10.1145/3308558.3313505}
Yufeng Diao, Hongfei Lin, Liang Yang, Xiaochao Fan, Di~Wu, Dongyu Zhang, and
  Kan Xu. 2019.
\newblock \href {https://doi.org/10.1145/3308558.3313505} {Heterographic pun
  recognition via pronunciation and spelling understanding gated attention
  network}.
\newblock In \emph{The World Wide Web Conference}, WWW '19, page 363–371, New
  York, NY, USA. Association for Computing Machinery.

\bibitem[{Garimella et~al.(2020)Garimella, Banea, Hossain, and
  Mihalcea}]{garimella-etal-2020-judge}
Aparna Garimella, Carmen Banea, Nabil Hossain, and Rada Mihalcea. 2020.
\newblock \href {https://doi.org/10.18653/v1/2020.coling-main.253} {{``}judge
  me by my size (noun), do you?{''} {Y}oda{L}ib: A demographic-aware humor
  generation framework}.
\newblock In \emph{Proceedings of the 28th International Conference on
  Computational Linguistics}, pages 2814--2825, Barcelona, Spain (Online).
  International Committee on Computational Linguistics.

\bibitem[{Ghazvininejad et~al.(2016)Ghazvininejad, Shi, Choi, and
  Knight}]{ghazvininejad2016generating}
Marjan Ghazvininejad, Xing Shi, Yejin Choi, and Kevin Knight. 2016.
\newblock Generating topical poetry.
\newblock In \emph{Proceedings of the 2016 Conference on Empirical Methods in
  Natural Language Processing}, pages 1183--1191.

\bibitem[{Goodfellow et~al.(2014)Goodfellow, Pouget-Abadie, Mirza, Xu,
  Warde-Farley, Ozair, Courville, and Bengio}]{goodfellow2014generative}
Ian Goodfellow, Jean Pouget-Abadie, Mehdi Mirza, Bing Xu, David Warde-Farley,
  Sherjil Ozair, Aaron Courville, and Yoshua Bengio. 2014.
\newblock Generative adversarial nets.
\newblock \emph{Advances in neural information processing systems}, 27.

\bibitem[{Hashimoto et~al.(2018)Hashimoto, Guu, Oren, and
  Liang}]{hashimoto2018retrieveandedit}
Tatsunori~B. Hashimoto, Kelvin Guu, Yonatan Oren, and Percy Liang. 2018.
\newblock \href {http://arxiv.org/abs/1812.01194} {A retrieve-and-edit
  framework for predicting structured outputs}.

\bibitem[{He et~al.(2019)He, Peng, and Liang}]{he2019pun}
He~He, Nanyun Peng, and Percy Liang. 2019.
\newblock Pun generation with surprise.
\newblock In \emph{2019 Conference of the North American Chapter of the
  Association for Computational Linguistics: Human Language Technologies, NAACL
  HLT 2019}, pages 1734--1744. Association for Computational Linguistics (ACL).

\bibitem[{Hong and Ong(2009)}]{hong-ong-2009-automatically}
Bryan~Anthony Hong and Ethel Ong. 2009.
\newblock \href {https://aclanthology.org/W09-2004} {Automatically extracting
  word relationships as templates for pun generation}.
\newblock In \emph{Proceedings of the Workshop on Computational Approaches to
  Linguistic Creativity}, pages 24--31, Boulder, Colorado. Association for
  Computational Linguistics.

\bibitem[{Kao et~al.(2016)Kao, Levy, and Goodman}]{kao2016computational}
Justine~T Kao, Roger Levy, and Noah~D Goodman. 2016.
\newblock A computational model of linguistic humor in puns.
\newblock \emph{Cognitive science}, 40(5):1270--1285.

\bibitem[{Li et~al.(2016)Li, Galley, Brockett, Gao, and
  Dolan}]{li2016diversity}
Jiwei Li, Michel Galley, Chris Brockett, Jianfeng Gao, and Bill Dolan. 2016.
\newblock A diversity-promoting objective function for neural conversation
  models.
\newblock In \emph{HLT-NAACL}.

\bibitem[{Lippman and Dunn(2000)}]{lippman2000contextual}
Louis~G Lippman and Mara~L Dunn. 2000.
\newblock Contextual connections within puns: Effects on perceived humor and
  memory.
\newblock \emph{The journal of general psychology}, 127(2):185--197.

\bibitem[{Luo et~al.(2019)Luo, Li, Yang, Li, Chang, Sui, and Sun}]{luo2019pun}
Fuli Luo, Shunyao Li, Pengcheng Yang, Lei Li, Baobao Chang, Zhifang Sui, and
  Xu~Sun. 2019.
\newblock Pun-gan: Generative adversarial network for pun generation.
\newblock In \emph{Proceedings of the 2019 Conference on Empirical Methods in
  Natural Language Processing and the 9th International Joint Conference on
  Natural Language Processing (EMNLP-IJCNLP)}, pages 3388--3393.

\bibitem[{Miller and Gurevych(2015)}]{miller-gurevych-2015-automatic}
Tristan Miller and Iryna Gurevych. 2015.
\newblock \href {https://doi.org/10.3115/v1/P15-1070} {Automatic disambiguation
  of {E}nglish puns}.
\newblock In \emph{Proceedings of the 53rd Annual Meeting of the Association
  for Computational Linguistics and the 7th International Joint Conference on
  Natural Language Processing (Volume 1: Long Papers)}, pages 719--729,
  Beijing, China. Association for Computational Linguistics.

\bibitem[{Miller et~al.(2017)Miller, Hempelmann, and
  Gurevych}]{miller2017semeval}
Tristan Miller, Christian~F Hempelmann, and Iryna Gurevych. 2017.
\newblock Semeval-2017 task 7: Detection and interpretation of english puns.
\newblock In \emph{Proceedings of the 11th International Workshop on Semantic
  Evaluation (SemEval-2017)}, pages 58--68.

\bibitem[{Mishra et~al.(2019)Mishra, Tater, and
  Sankaranarayanan}]{mishra2019modular}
Abhijit Mishra, Tarun Tater, and Karthik Sankaranarayanan. 2019.
\newblock A modular architecture for unsupervised sarcasm generation.
\newblock In \emph{Proceedings of the 2019 Conference on Empirical Methods in
  Natural Language Processing and the 9th International Joint Conference on
  Natural Language Processing (EMNLP-IJCNLP)}, pages 6144--6154.

\bibitem[{Mou et~al.(2015)Mou, Yan, Li, Zhang, and Jin}]{mou2015backward}
Lili Mou, Rui Yan, Ge~Li, Lu~Zhang, and Zhi Jin. 2015.
\newblock Backward and forward language modeling for constrained sentence
  generation.
\newblock \emph{arXiv preprint arXiv:1512.06612}.

\bibitem[{Petrovi{\'c} and
  Matthews(2013)}]{petrovic-matthews-2013-unsupervised}
Sa{\v{s}}a Petrovi{\'c} and David Matthews. 2013.
\newblock \href {https://aclanthology.org/P13-2041} {Unsupervised joke
  generation from big data}.
\newblock In \emph{Proceedings of the 51st Annual Meeting of the Association
  for Computational Linguistics (Volume 2: Short Papers)}, pages 228--232,
  Sofia, Bulgaria. Association for Computational Linguistics.

\bibitem[{Qi et~al.(2020)Qi, Zhang, Yang, Liu, and Sun}]{qi2020wantwords}
Fanchao Qi, Lei Zhang, Yanhui Yang, Zhiyuan Liu, and Maosong Sun. 2020.
\newblock Wantwords: An open-source online reverse dictionary system.
\newblock In \emph{Proceedings of the 2020 Conference on Empirical Methods in
  Natural Language Processing: System Demonstrations}, pages 175--181.

\bibitem[{Raffel et~al.(2020)Raffel, Shazeer, Roberts, Lee, Narang, Matena,
  Zhou, Li, and Liu}]{raffel2020exploring}
Colin Raffel, Noam Shazeer, Adam Roberts, Katherine Lee, Sharan Narang, Michael
  Matena, Yanqi Zhou, Wei Li, and Peter~J Liu. 2020.
\newblock Exploring the limits of transfer learning with a unified text-to-text
  transformer.
\newblock \emph{Journal of Machine Learning Research}, 21:1--67.

\bibitem[{Rose et~al.(2010)Rose, Engel, Cramer, and Cowley}]{rose2010automatic}
Stuart Rose, Dave Engel, Nick Cramer, and Wendy Cowley. 2010.
\newblock Automatic keyword extraction from individual documents.
\newblock \emph{Text mining: applications and theory}, 1:1--20.

\bibitem[{Shahaf et~al.(2015)Shahaf, Horvitz, and Mankoff}]{shahaf2015inside}
Dafna Shahaf, Eric Horvitz, and Robert Mankoff. 2015.
\newblock Inside jokes: Identifying humorous cartoon captions.
\newblock In \emph{Proceedings of the 21th ACM SIGKDD International Conference
  on Knowledge Discovery and Data Mining}, pages 1065--1074.

\bibitem[{Stowe et~al.(2021)Stowe, Chakrabarty, Peng, Muresan, and
  Gurevych}]{stowe2021metaphor}
Kevin Stowe, Tuhin Chakrabarty, Nanyun Peng, Smaranda Muresan, and Iryna
  Gurevych. 2021.
\newblock \href {http://arxiv.org/abs/2106.01228} {Metaphor generation with
  conceptual mappings}.

\bibitem[{Tian et~al.(2021)Tian, Sridhar, and
  Peng}]{tian-etal-2021-hypogen-hyperbole}
Yufei Tian, Arvind~krishna Sridhar, and Nanyun Peng. 2021.
\newblock \href {https://aclanthology.org/2021.findings-emnlp.136}
  {{H}ypo{G}en: Hyperbole generation with commonsense and counterfactual
  knowledge}.
\newblock In \emph{Findings of the Association for Computational Linguistics:
  EMNLP 2021}, pages 1583--1593, Punta Cana, Dominican Republic. Association
  for Computational Linguistics.

\bibitem[{Valitutti et~al.(2013)Valitutti, Toivonen, Doucet, and
  Toivanen}]{inproceedings2000}
Alessandro Valitutti, Hannu Toivonen, Antoine Doucet, and Jukka Toivanen. 2013.
\newblock "let everything turn well in your wife": Generation of adult humor
  using lexical constraints.
\newblock volume~2.

\bibitem[{Yang et~al.(2020)Yang, Cui, Chen, Che, Liu, Wang, and
  Hu}]{yang-etal-2020-textbrewer}
Ziqing Yang, Yiming Cui, Zhipeng Chen, Wanxiang Che, Ting Liu, Shijin Wang, and
  Guoping Hu. 2020.
\newblock \href {https://doi.org/10.18653/v1/2020.acl-demos.2} {{T}ext{B}rewer:
  {A}n {O}pen-{S}ource {K}nowledge {D}istillation {T}oolkit for {N}atural
  {L}anguage {P}rocessing}.
\newblock In \emph{Proceedings of the 58th Annual Meeting of the Association
  for Computational Linguistics: System Demonstrations}, pages 9--16, Online.
  Association for Computational Linguistics.

\bibitem[{Yu et~al.(2018)Yu, Tan, and Wan}]{yu2018neural}
Zhiwei Yu, Jiwei Tan, and Xiaojun Wan. 2018.
\newblock A neural approach to pun generation.
\newblock In \emph{Proceedings of the 56th Annual Meeting of the Association
  for Computational Linguistics (Volume 1: Long Papers)}, pages 1650--1660.

\bibitem[{Yu and Wan(2019)}]{yu-wan-2019-avoid}
Zhiwei Yu and Xiaojun Wan. 2019.
\newblock \href {https://doi.org/10.18653/v1/N19-1092} {How to avoid sentences
  spelling boring? towards a neural approach to unsupervised metaphor
  generation}.
\newblock In \emph{Proceedings of the 2019 Conference of the North {A}merican
  Chapter of the Association for Computational Linguistics: Human Language
  Technologies, Volume 1 (Long and Short Papers)}, pages 861--871, Minneapolis,
  Minnesota. Association for Computational Linguistics.

\bibitem[{Yu et~al.(2020)Yu, Zang, and Wan}]{yu-etal-2020-homophonic}
Zhiwei Yu, Hongyu Zang, and Xiaojun Wan. 2020.
\newblock \href {https://doi.org/10.18653/v1/2020.emnlp-main.229} {Homophonic
  pun generation with lexically constrained rewriting}.
\newblock In \emph{Proceedings of the 2020 Conference on Empirical Methods in
  Natural Language Processing (EMNLP)}, pages 2870--2876, Online. Association
  for Computational Linguistics.

\bibitem[{Zhang et~al.(2020)Zhang, Qi, Liu, Wang, Liu, and
  Sun}]{zhang2020multi}
Lei Zhang, Fanchao Qi, Zhiyuan Liu, Yasheng Wang, Qun Liu, and Maosong Sun.
  2020.
\newblock Multi-channel reverse dictionary model.
\newblock In \emph{Proceedings of the AAAI Conference on Artificial
  Intelligence}, pages 312--319.

\bibitem[{Zou and Lu(2019)}]{zou-lu-2019-joint}
Yanyan Zou and Wei Lu. 2019.
\newblock \href {https://doi.org/10.18653/v1/N19-1217} {Joint detection and
  location of {E}nglish puns}.
\newblock In \emph{Proceedings of the 2019 Conference of the North {A}merican
  Chapter of the Association for Computational Linguistics: Human Language
  Technologies, Volume 1 (Long and Short Papers)}, pages 2117--2123,
  Minneapolis, Minnesota. Association for Computational Linguistics.

\end{thebibliography}
\bibliographystyle{acl_natbib}

\cleardoublepage
\appendix
 \begin{table*}[t]
\centering
\small
\begin{tabular}{@{}cc| c c@{}}
\toprule
\multicolumn{2}{l}{} &
  \textbf{Sense 1} &
  \textbf{Sense 2} \\ \toprule
\multicolumn{2}{c}{ \textbf{Definition}} &
  \begin{tabular}[c]{@{}c@{}}a string of words satisfying\\  the grammatical rules of a language\end{tabular} &
  \begin{tabular}[c]{@{}c@{}}a final judgment of guilty in a\\  criminal case and the punishment\\  that is imposed\end{tabular} \\ \hline
\multicolumn{2}{c}{\textbf{Related words}} &
  \begin{tabular}[c]{@{}c@{}}syllable, syntax, lexicon, thesaurus,\\  grammatical\end{tabular} &
  \begin{tabular}[c]{@{}c@{}}punishment, verdict, sentencing, \\ retrial, penalty\end{tabular} \\ \toprule
\multicolumn{2}{c}{\textbf{TF-IDF}} &

  \begin{tabular}[c]{@{}c@{}} syllables, words, three, spelling, \\ even, said, describe, typos\end{tabular} &
  \begin{tabular}[c]{@{}c@{}} cruel, expected, end, court,\\  scheduled, set, spector, seeking\\
  \end{tabular} \\ \hline 
\multicolumn{2}{c}{\textbf{Word2Vec}} &
\begin{tabular}[c]{@{}c@{}} syllable, pronounced, words, rhyme,\\ verbs, meaning, hence, example \end{tabular} &
  \begin{tabular} [c]{@{}c@{}}punished, crimes, offender, torture, \\ moral, guilt, abuse, offender \\
  \end{tabular}\\ \hline 
\multicolumn{2}{c}{\textbf{GPT3}} &
  \begin{tabular}[c]{@{}c@{}}words, letters, punctuation, grammar,\\ synonym, dictionary, meaning, comma 
  \end{tabular} &
  \begin{tabular}[c]{@{}c@{}}prison, judge, jury, trial,\\  justice, lawyer, court, evidence\end{tabular} \\ \bottomrule
\end{tabular}%
\caption{Comparison of the three different context word generation mechanism for the pun word `sentence'. The table lists two sense definitions and the related words obtained from the sense definitions using reverse dictionary. For these related words, we obtain context words using three different mechanisms.}
\label{table: context words}
\end{table*}
\section*{Appendix}

\section{Details in Experiments}
\subsection{An Example of Context Words}
We list the output of context words for the pun word `sentence' in Table \ref{table: context words}. The table lists two sense definitions and the related words obtained from the sense definitions using reverse dictionary. We then obtain context words using three different mechanisms: TF-IDF, Word2Vec, and GPT3.

\subsection{Implementation Details}\label{subsec:implement}
\paragraph{Experimental Settings}
 For the word2vec model we train a continuous-bag-of-words model with window size 40 and word vector dimension 200. For the candidate generation module, we train the T5-base model on 10 epochs and select the best performing model based on validation loss.  Max sequence length for target and source is set to 30. Batch size is set to 64. 

\paragraph{Data Refinement} The process to generate  both related and context words can entail many words that are not ideal. Continuing with these words would further propagate and enlarge the noise. Hence, to minimize this noise, we implement the following data refinement steps to ensure the keywords stick to our standards: we avoid using polysemous words as keywords during intermediate steps because their perceived sense is highly ambiguous. We also disregard any numbers and special characters produced by our systems.

\subsection{Human Evaluation}
The workers are paid \$20 per hour. For pun success judgement (yes/no question), we take the majority vote among three workers, while for funniness and coherence (1 to 5), we take the average ratings. We then use the pairwise kappa coefficient to measure the inter-annotator agreement (IAA). The average inter-annotator agreement of all raters for pun success, funniness and coherence are 0.55, 0.48 and 0.40, meaning that annotators moderately agree with each other. Considering the subjectivity of this task \cite{braslavski2018evaluate}, and the higher IAA in terms of pun success and funniness over coherence, we argue that our collected results are reasonably reliable for the purpose of pun generation. Besides, we conducted paired t-test and show that the success rate and funniness ratings of our systems differentiate from the best baseline model with statistical significance (p-value < 0.5).
 
\section{Humor Classifier Results for Selecting Puns}\label{appendix:humor-classification}

To further discuss the accuracy and recall of our humor classifier, we show a representative output in Table \ref{table:classifier}. The table contains a few selected sentences ranked my the humor classifier. We also label each sentence as \textit{yes}, \textit{no}, and \textit{maybe} to indicate if it is a pun or not. As discussed in the methodology, we train our classifier on humor dataset. As puns are an important part of humor generation, this model can help rule out some options. Basic theories of humor such as incongruity and surprise apply to both of them. As can be seen in the table, our classifier is able to successfully pull aside unfunny or non-coherent sentences. Looking at the examples at the top and the middle, it can be observed that some better examples are classified lower than others. Making this observation across many pun words, we decided to use the classifier only to rule out the bottom third samples. For the rest of the generations, we randomly sample them. 

On manual observation, we realised that when we as humans peruse the generated candidates, there are many sentences that meet our expectations. Therefore, building a classifier that can accurately find these sentences can increase the accuracy by a large margin. We treat this as an opportunity for future work.

\begin{table*}[t]
\centering
\small
\begin{tabular}{@{}lcc@{}}
\toprule
Sentence                                                                                          & Rank & Pun   \\ \midrule
What's the interest rate on a home mortgage? No interest.                 & 1    & Yes   \\
My bank said I think they're interested in me. I said no.                 & 2    & No    \\
My girlfriend said she had an interest in banking so i loan her a quarter & 3    & Yes   \\
I have no interest in being a guardian. It's free.                        & 4    & Maybe \\
I've never had interest placed on borrowings. It's a waste of time.       & 5    & Yes   \\
Why did the republican attack the bank? Because it was in its interest.   & 6    & Maybe \\
What is the republican's strategy? The interest rate.                     & 7    & No    \\
What is the most dispensable interest in investment?                      & 8    & No    \\
If trump had an interest in president he would make it an president-of-interest. & 9 & No \\ \bottomrule
\end{tabular}
\caption{An example of candidate pun sentences ranked by the humor classifier. As can be seen, the model is able to rule out non-pun sentences but fails to pick out high-quality ones.}
\label{table:classifier}
\end{table*}

\begin{table}[t]
\centering
\small
\begin{tabular}{@{}ll@{}}
\toprule
Target word & \textbf{sentence}                                                                \\ \midrule
Sense 1     & A string of words satisfying the grammatical rules of a language        \\
Sense 2 & (Criminal law) a final judgment of guilty in a criminal case and the punishment that is imposed \\ \midrule
1           & The word jail is a sentence.                                            \\
2       & What's the punishment for using antonyms in a sentence syntax is it a sentence?                 \\
3           & I'm sorry I said the sentence was too long but punishments are endless. \\
4           & The sentence in the dictionary doesn't sound very guilty.               \\ \bottomrule

\\\toprule
Target word & \textbf{case}                                              \\
Sense 1     & A portable container for carrying several objects          \\
Sense 2 & A statement of facts and reasons used to support an argument                    \\ \midrule
1           & What's the most durable luggage for a detective? jury case \\
2       & A jury just found a container of leather there's no reason to argue it's a case \\
3       & What do you call a container used for investigation research? a case study      \\
4       & Why did the cardboard get into a trial? because it was an investigation case  \\ \bottomrule
\\ \toprule
Target word & \textbf{bugs}                                                                     \\ \midrule
Sense 1     & General term for any insect or similar creeping or crawling invertebrate \\
Sense 2     & A fault or defect in a computer program, system, or machine              \\ \midrule
1           & Why did the garden restart its computer? it had bugs in it.              \\
2           & What do you call a pest that's slow programmer? bugs bug                 \\
3           & Why did the compost crash? it had bugs in it.                            \\
4           & What do you call a bug that's disgusting? a glitch in the internet       \\ \bottomrule

\\ \toprule
Target word & \textbf{delivery}                                                                     \\ \midrule
Sense 1     & the act of delivering or distributing something (as goods or mail) \\
Sense 2     & your characteristic style or manner of expressing yourself orally              \\ \midrule
1           & What did the letter say to the parcel? clear delivery!              \\
2           & What do you call a trucking truckdriver with no articulation? delivery driver.                 \\
3           & The distribution center has a pronunciation dictionary. it's a delivery service                            \\
4           & What do you call a parcel with no dialogue and an accent? delivery service.       \\ \bottomrule

\end{tabular}

\caption{More examples generated by \ModelName.}
\label{table:more-examples}
\end{table}

\section{Analysis of Human Written Puns}
we calculate the position of the pun words of 1,163 human written
pun sentences from SemEval 2017 Task 7 and report the distribution.
The histogram corroborates with the human annotated samples in that both suggest that keeping the
pun word at the end of the sentence generates funnier puns. Theory of humor which says that the
``joke'' in a funny sentences some towards the end
of the sentence validates our analysis.
\begin{figure}[t]
  \centering
    \includegraphics[width=0.35\textwidth]{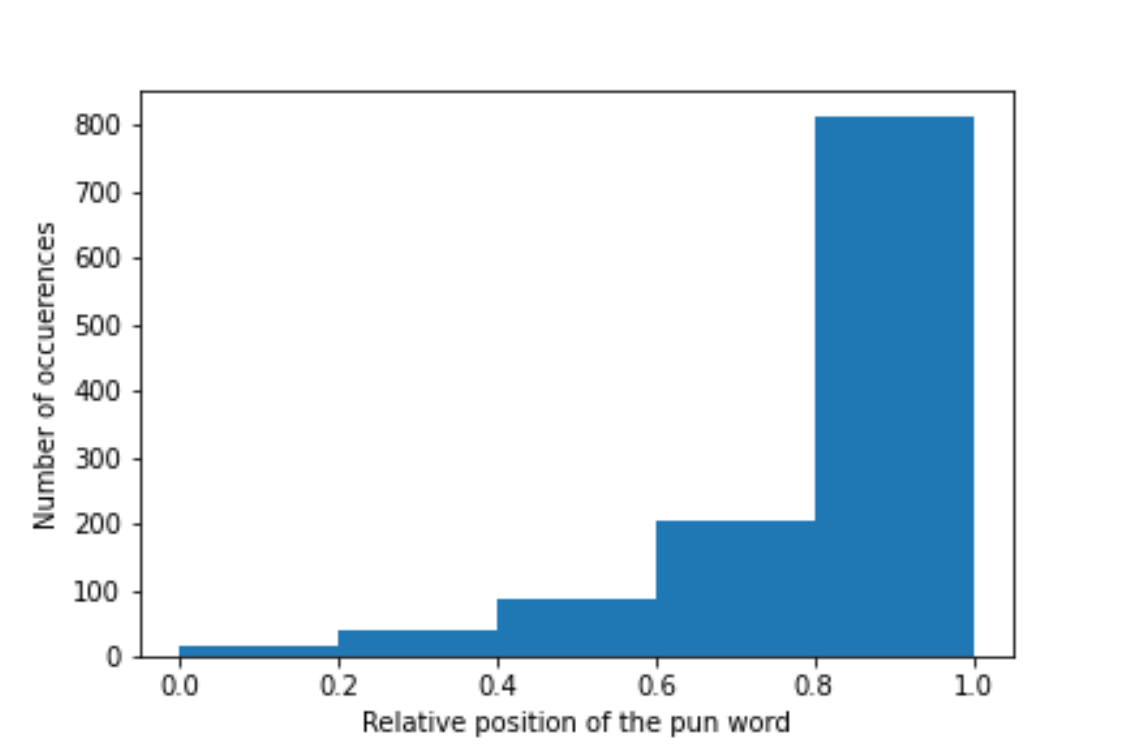}
  \caption{Analysis of the position of pun word in 1,163 human written puns. The  y-axis indicates the number of sentences and the x-axis indicates the position of pun word on a scale from 0 (start) to 1 (end).} 
  \label{fig: position}    
  \vspace{-1.5em}
\end{figure}

\section{More Examples of Generated Puns}\label{appendix:examples}
We compile more examples generated by \BaseName in Table \ref{table:more-examples} for the following pun words: \textit{sentence}, \textit{case}, \textit{bugs}, \textit{delivery}. This table further supports our claim that our approach would benefit from a better classification module to select human-like sentences. 

 \section{GPT3 for context words generation} \label{sec:gpt3 appendix}

We make use of few shot GPT3 to generate context words. The prompy to GPT3 included 2 pair of prompt and its completion. One example of a pair would be \textit{``generate seven keywords for laptop: battery, macbook, internet, technology, keyboard, technology, portable''}. These example are followed by the prompt \textit{``generate seven keywords for X:''} where X is  a related word. This way we generate seven keywords for each related word.

\end{document}